\documentclass[letterpaper, 10 pt, conference]{ieeeconf}  

\IEEEoverridecommandlockouts                              
                                                          
\usepackage{multirow}

\usepackage{url}
\usepackage{amsmath}
\usepackage{cite}
\usepackage[dvips]{graphicx}
\usepackage{epsfig}
\usepackage{epic,eepic,color}
\usepackage{times}
\usepackage{mathptmx}

\usepackage{amsmath}
\usepackage{amssymb}

\usepackage{booktabs}

\usepackage{cases}

\usepackage{tabularx}

\usepackage{times}

\definecolor{red}{rgb}{1,0,0}
\definecolor{green}{rgb}{0,1,0}
\definecolor{blue}{rgb}{0,0,1}
\definecolor{violet}{rgb}{1,0,1}
\definecolor{cyan}{cmyk}{1,0,0,0}
\definecolor{magenta}{cmyk}{0,1,0,0}
\definecolor{yellow}{cmyk}{0,0,1,0}

\definecolor{white}{rgb}{1,1,1}
\definecolor{PrimaryGreen}{rgb}{0.10,0.45,0.20}

\newcommand{\CommentOut}[1]{}

\onecolumn

\begin{document}

\title{%
Textual Scene Mapper:\\ Continual Scene Change Identification Under Image Disposal
} 

\author{Phalguna Reddy Baddam ~~~~~~~~~~~~~ Kanji Tanaka~~~~~~~~~~~~~
\thanks{This work was partially supported by MECSST Grant (23K11270, 26K15002), by KURATA grants and by TATEISI Science And Technology Foundation.}
\thanks{P. R. Baddam and K.Tanaka are with Graduate School of Engineering, University of Fukui, Japan.
        {\tt\small tnkknj@u-fukui.ac.jp}}%
}

\maketitle

\begin{abstract}
For mobile robots performing long-term autonomy, maintaining a consistent environment map requires continuously tracking dynamic state changes (Scene Change Identification: SCI). Textual Scene Memory (TSM), which discards high-resolution raw images and retains only lightweight textual representations, offers storage efficiency and semantic generalization. Rather than treating text as passive logs, we formulate TSM as explicit linguistic scene parameterization, where structured-and-descriptive text serves as symbolic model parameters queried by a frozen foundation model under unknown future state-change queries. However, reliable linguistic parameterization under strict raw-image disposal faces three challenges: (1) localization-error sensitivity from naive coordinate string serialization, (2) false positives from lexical drift across viewpoint shifts, and (3) catastrophic forgetting--viewed through replay-free continual learning--when temporary occlusions are misread as physical disappearance and overwrite accumulated parameters. We formulate long-term TSM maintenance as a Spatio-Temporal Decentralized Partially Observable Markov Decision Process (Dec-POMDP) and establish the formal requirements specification and pluggable architecture for Textual Scene Mapper (TSM). We derive requirements (REQ-1.1-REQ-3.4) across qualitative spatial abstraction, hybrid textual encoding ($o_t = o_t^{free} \oplus o_t^{struct}$) with static landmarks and unknown occlusion tags, Bayesian surprise gating, change memory pools, context retrieval, occlusion gating, and two-stage offline edge policy adaptation with deterministic verifiers. Proof-of-Concept validation on 10 multi-view sets in Isaac Sim 5.1.0 raises recall from 90\% ($9/10$) to 100\% ($10/10$) and cuts false positives by 72.2\% (255 to 71) over the free-form baseline.
\end{abstract}

\section{Introduction}
\label{sec:introduction}

\subsection{Background and Motivation: Necessity of Textual Scene Memory in Long-Term Autonomy}
\label{sec:motivation}
To execute long-term autonomy over weeks or months in indoor and human-inhabited environments, a mobile robot must continuously track dynamic state changes caused by human activities and object relocations (Scene Change Identification: SCI) and maintain an internally consistent map. Conventional 3D geometric maps (e.g., point clouds and occupancy grids) and high-dimensional local invariant feature maps suffer from storage bloat during large-scale, long-term operation, degraded distinctiveness in environments with repetitive structures (the distinctiveness--invariance trade-off), and sensitivity to illumination variations.

In contrast, Textual Scene Memory (TSM)---which discards high-resolution raw sensor data from local spatial observations and converts spatial structures and semantic information into compact symbolic and linguistic descriptions (Map-to-Text)---is a promising paradigm that achieves fast matching and high semantic generalization with a minimal memory footprint. Crucially, whereas conventional approaches parameterize an environment via 3D geometric coordinates or continuous neural network weights updated through gradient descent, TSM treats the structured-and-descriptive text itself as \emph{explicit linguistic scene parameters} interpreted by a frozen foundation model. Unlike dense visual embeddings or neural scene representations (e.g., 3D Gaussian Splatting) that remain tightly coupled to specific encoder architectures and prohibit direct human inspection, text serves as a model-agnostic, human- and LLM-interpretable persistent medium that remains compatible across foundation model upgrades while structurally filtering out personally identifiable information (PII) via predefined schemas. Symbolizing spatial layouts using unique local viewpoints or static landmarks as reference anchors yields significantly higher spatial discriminability than unordered feature sets. However, extending this paradigm from single-shot place recognition to sequential memory maintenance and updating under continuous environmental dynamics requires a rigorous, implementation-agnostic system specification.

\subsection{Boundary Between Geometric and Textual Representations: Qualitative Spatial Abstraction}
\label{sec:qualitative_principle}
When specifying the requirements for long-term TSM, it is first necessary to establish a clear boundary between coordinate data in conventional geometric SLAM and the textual scene memory studied in this work. Conventional geometric SLAM records 3D point clouds and object poses as millimeter-precision floating-point coordinates $[x, y, z]$. Simply converting these numbers into ASCII strings constitutes mere \emph{coordinate serialization} rather than symbolic or linguistic spatial memory. Raw coordinate strings fluctuate every time due to minor localization errors or viewpoint offsets, causing temporal string comparisons to break down while preventing language models (LLMs) from leveraging commonsense and spatial contextual reasoning.

Therefore, establishing genuine textual scene memory requires the fundamental architectural requirement of \emph{eliminating high-precision metric coordinate lists and abstracting spatial information into qualitative, intuitive descriptions that humans and language models can semantically reason about}. As one illustrative example of a design guideline that realizes this boundary, spatial distances, areas, and scales can be quantized into approximate expressions within two significant digits (e.g., ``a space of about 1.2 tatami mats,'' ``a corridor approximately 35~ft ahead,'' or ``right next to the shelf''), matching the granularity at which humans perceive space in daily conversation. By separating precise metric measurement (metric SLAM) from qualitative, semantic long-term memory (TSM), we define a linguistic parameter basis that is directly amenable to natural language reasoning in the shared semantic space of LLMs without geometric reconstruction or model weight updates.

\subsection{Technical Challenges Requiring Architectural Specification}
\label{sec:challenges}
Unlike conventional continual learning in continuous neural weight spaces with experience replay, maintaining an environment map solely as discrete linguistic parameters under strict raw-image disposal and absent deployment-time supervision requires overcoming three structural challenges:
\begin{enumerate}
    \item \textbf{Linguistic Parameterization Under Unknown Future Queries and Lexical Drift:} Because raw visual observations are immediately discarded at memory encoding time, the front-end must write down a textual parameter representation before knowing which specific object state or spatial relation will change in the future. However, when visual observations are converted directly into naive free-form natural language text, variations in the robot's patrol trajectory or camera viewpoint ($v_A \rightarrow v_B$) induce lexical drift even when the physical environment remains completely static. Consequently, temporal text comparisons suffer from frequent false positives, misidentifying unchanged objects as changed.
    \item \textbf{Catastrophic Forgetting Due to Occlusion (Continual Learning Analogy):} Through the analogy of continual learning in SLAM, the irreversible translation from raw visual scenes into compact textual parameters under immediate image disposal can be viewed directly as learning model parameters from sensory experience without experience replay. Under this analogy, when a known object temporarily leaves the camera's field of view due to viewpoint shifts or foreground obstacles, mistaking this temporary occlusion for physical disappearance and deleting or overwriting existing textual records manifests as \emph{catastrophic forgetting}, irreversibly destroying previously accumulated symbolic parameters. Because raw images cannot be rehearsed post hoc, preventing catastrophic forgetting in replay-free linguistic memory remains a fundamental open challenge---and a central target for our continual learning extensions---requiring a structural mechanism that strictly decouples extrinsic observation noise (viewpoint shifts and temporary occlusions) from intrinsic environmental changes.
    \item \textbf{Limitations of Monolithic Design and Lack of Verifiability:} Integrating local text encoding, event gating, spatio-temporal context retrieval, uncertainty filtering, and edge-model adaptation into a single monolithic black box makes it impossible to isolate which stage caused a memory corruption error. A formal modular specification with strict input--output interfaces is required so that core representations can be validated independently before scaling to full programmatic automation and on-device edge deployment without prematurely locking in specific underlying algorithms.
\end{enumerate}

\subsection{Our Approach: Requirements Specification and Phased Validation Scope}
\label{sec:approach}
To address these challenges without restricting future algorithmic choices, we formulate long-term textual scene memory maintenance as a Spatio-Temporal Decentralized Partially Observable Markov Decision Process (Dec-POMDP)~\cite{oliehoek2016concise} and establish the system requirements specification for \emph{Textual Scene Mapper (TSM)}, a pluggable functional architecture composed of model- and algorithm-agnostic functional blocks (\texttt{REQ-1.1}--\texttt{REQ-3.4}).

At the core of this specification is an explicit linguistic parameter representation at encoding time that generalizes across unknown future change queries without retaining raw images. Under the qualitative spatial abstraction principle, rather than compressing visual information into a single unstructured textual modality, the front-end specifies two parallel encoders: a \emph{free-form description encoder} ($o_t^{\text{free}}$) that flexibly verbalizes spatial layouts and visual attributes, and a \emph{canonical state encoder} ($o_t^{\text{struct}}$) that extracts discrete states anchored to invariant spatial landmarks alongside explicit uncertainty tags (\texttt{unknown}) representing out-of-view or occluded regions based on a canonical vocabulary schema. These outputs are fused via \emph{hybrid textual encoding} ($o_t = o_t^{\text{free}} \oplus o_t^{\text{struct}}$). Grounding relative spatial descriptions and discrete state schemas in static landmarks preserves compatibility with future geometric viewpoint planning and local map descriptors while achieving viewpoint invariance and preventing memory corruption under occlusion~\cite{ahn2022saycan}.

Crucially, to maintain rigorous academic transparency while preserving engineering flexibility, this paper clearly distinguishes between \emph{architectural requirements specification} and \emph{current empirical validation status}:
\begin{itemize}
    \item \textbf{Complete System Requirements Specification (All Modules):} We specify the functional requirements, mathematical formulations, input--output contracts, and theoretical rationale (\emph{why} each mechanism is required to prevent failure modes) for all modules across the full long-horizon architecture---including representation-space Bayesian surprise observation gating, self-evolving change memory pools, spatio-temporal context retrieval, uncertainty filtering, and two-stage offline policy adaptation for compact on-device language models guided by deterministic rule-based verifiers. All extension modules are specified purely in terms of functional invariants and interface contracts rather than specific third-party libraries, model families, or optimization implementations.
    \item \textbf{Foundational Proof-of-Concept (PoC) Validation (Core Representation):} At the current initial stage, we empirically validate the foundational core of this specification---namely, hybrid textual encoding ($o_t^{\text{free}} \oplus o_t^{\text{struct}}$) and dialogue-style state-change reasoning---under a strictly controlled manual PoC protocol in a 3D physics simulator, isolating representation invariance from downstream system-level noise before executing programmatic automation.
\end{itemize}

\subsection{Main Contributions}
\label{sec:contributions}
The main contributions of this paper are summarized as follows:
\begin{enumerate}
    \item \textbf{Formal Problem Formulation as a Spatio-Temporal Dec-POMDP (Sections~\ref{sec:formulation}--\ref{sec:taxonomy}):} We formalize long-term map maintenance under strict raw-image disposal as a Spatio-Temporal Dec-POMDP in which hybrid text serves as explicit scene model parameters encoded prior to knowing future change queries. Under the qualitative spatial abstraction principle (e.g., via two-significant-digit approximations), we systematize six categories of intrinsic scene changes distinct from extrinsic observation noise and specify a discrete memory action space ($\mathcal{A}_{\text{mem}}$).
    \item \textbf{Complete Modular Requirements Specification and Rationale (Sections~\ref{sec:architecture}--\ref{sec:extension_specs}):} We define strict functional requirements (\texttt{REQ-1.1}--\texttt{REQ-3.4}) and input--output interfaces for all pipeline modules, deriving the theoretical and information-theoretic rationale explaining why each component (hybrid encoding, \texttt{unknown} uncertainty gating, Bayesian surprise triggers, self-evolving memory pools, and two-stage offline edge policy adaptation with deterministic verifiers) is necessary for long-horizon consistency without binding the architecture to specific existing implementations.
    \item \textbf{Foundational PoC Validation of Core Representation Requirements (Section~\ref{sec:experiments}):} Through controlled multi-view experiments in Isaac Sim 5.1.0 (Sets~1--10) that isolate representation performance from system-level confounding factors, we empirically confirm that our hybrid representation achieves 100\% recall (10/10) of true state changes while reducing false positives by 72.2\% compared to free-form text alone, and we establish quantitative metrics ($\text{MC-F1}$, $\text{CFR}$, and $\text{IAR}$) for benchmarking subsequent automated phases.
\end{enumerate}

\subsection{Paper Organization}
\label{sec:organization}
Section~\ref{sec:related_work} reviews related work and clarifies the positioning of our framework. Section~\ref{sec:method} presents the mathematical formulation, change taxonomy, complete functional requirements specification (\texttt{REQ-1.1}--\texttt{REQ-3.4}), and formal mathematical and information-theoretic specifications for every module. Section~\ref{sec:experiments} presents the controlled PoC validation of the core encoding and reasoning modules, including the full dataset visualizations, per-set results, and methodological justification. Section~\ref{sec:conclusion} concludes the paper and outlines the implementation roadmap.

\section{Related Work}
\label{sec:related_work}

This section reviews four primary research areas relevant to maintaining Textual Scene Memory (TSM) in long-term autonomy, highlighting the limitations of existing methods and clarifying the academic positioning of our modular requirements specification.

\subsection{Scene Change Identification in Long-Term Autonomy}
\label{sec:rw_sci}
Research on Scene Change Identification (SCI) for long-term autonomy has traditionally evolved around 3D point cloud differencing, occupancy grid map updates, and outlier rejection in Visual Place Recognition (VPR). Geometric change detection methods achieve high spatial precision by aligning millimeter-resolution point clouds acquired from LiDAR or depth sensors. However, they suffer from severe storage bloat over extended operation and frequent false positives caused by minor localization drift or sensor noise.

Furthermore, in appearance-based and deep-feature scene change detection, methods typically assume simultaneous access to paired images or stored visual embeddings, and it remains difficult to decouple extrinsic domain variations---such as illumination shifts and camera viewpoint changes ($v_A \rightarrow v_B$)---from intrinsic environmental changes like furniture relocation or door opening and closing. Unlike conventional metric- or feature-based SCI approaches that rely on geometric parameters or stored visual representations, our specification enforces immediate disposal of raw images and visual embeddings at observation time, performing change identification directly over explicit linguistic scene parameters abstracted into qualitative, intuitive spatial descriptions.

\subsection{Symbolic Spatial Representations and Textual Scene Memory}
\label{sec:rw_tsm}
To compactly represent environmental structures under strict computational and bandwidth constraints, converting local spatial observations or 3D scene graphs into symbolic and linguistic descriptions has attracted increasing attention. Classical symbolization methods based on visual Bag-of-Words or unordered local invariant features enable fast matching, yet they discard spatial layouts and inter-object topological relationships, resulting in degraded discriminability in indoor environments with repetitive structures. Conversely, methods that anchor spatial configurations to a canonical reference viewpoint or invariant landmarks achieve strong discriminability, while recent object-state benchmarks such as OSCaR~\cite{nguyen2024oscar} and ChangeIt-Frames~\cite{soucek2022look} demonstrate the utility of textual and discrete state vocabularies for describing object states. Nevertheless, most existing efforts focus on static single-shot place recognition, visual question answering conditioned on a known query, or isolated clip-level state classification rather than specifying persistent scene parameter representations capable of answering unknown future state-change queries under continuous environmental dynamics.

Moreover, naive Map-to-Text approaches that prompt Vision-Language Models (VLMs) to generate free-form captions directly from images are optimized for describing current visual appearance rather than serving as invariant scene parameters; consequently, they suffer from lexical drift across different viewpoints of an identical static scene, or degenerate into raw coordinate serialization by storing floating-point coordinates as plain text strings. Under our qualitative spatial abstraction principle (e.g., two-significant-digit approximations as one illustrative guideline), we specify fusing a free-form description encoder ($o_t^{\text{free}}$) with a static-landmark-anchored discrete state schema ($o_t^{\text{struct}}$) via hybrid textual encoding, constructing an explicit linguistic parameter representation that achieves both descriptive expressiveness and viewpoint invariance without locking into a single external ontology.

\subsection{Catastrophic Forgetting and Occlusion Robustness in Continual Learning}
\label{sec:rw_forgetting}
Systems that incrementally accumulate knowledge in dynamic environments face the stability--plasticity dilemma in continual learning. Viewed through the continual learning analogy in SLAM---where irreversible scene-to-text translation under immediate image disposal corresponds to updating symbolic scene parameters from sensory experience without experience replay---maintaining textual scene parameters introduces four distinct challenges beyond conventional neural network adaptation: (1) an \emph{irreversible, replay-free information bottleneck}, where discarding raw images prohibits experience replay and requires encoding query-agnostic sufficient statistics in a single pass; (2) \emph{non-stationary environmental dynamics versus observation shifts}, requiring the system to distinguish genuine physical changes (demanding plasticity) from viewpoint offsets or temporary occlusions (demanding stability) solely in text space; (3) \emph{discrete symbolic parameter editing} rather than smooth gradient-based weight regularization, where unconstrained text rewriting induces lexical drift and breaks entity identity; and (4) \emph{delayed or absent supervision at deployment}, where no ground-truth feedback indicates whether a missing entity has physically vanished or merely left the field of view. In particular, mistaking temporary occlusion for physical disappearance triggers severe \emph{catastrophic forgetting} via the erroneous deletion or overwriting of valid historical records.

Conventional approaches based on Bayesian state filtering or experience replay rely on continuous existence probability updates or raw data rehearsal, making them unsuitable under strict memory constraints where raw images are immediately discarded. By combining hybrid textual encoding ($o_t^{\text{free}} \oplus o_t^{\text{struct}}$) with an explicit uncertainty tag (\texttt{unknown}) representing unobservable or occluded states and enforcing a structural decoupling rule between \texttt{known} $\rightarrow$ \texttt{unknown} transitions and physical removal (Section~\ref{sec:architecture}), our specification resolves the discrete stability--plasticity dilemma and prevents catastrophic forgetting directly within the linguistic parameter space without retaining raw images.

\subsection{Sequential Decision-Making and Modular Memory Management with Language Models}
\label{sec:rw_llm}
Recent studies have rapidly advanced the use of Large Language Models (LLMs) and Small Language Models (SLMs) for robotic decision-making and external memory control via context retrieval. However, numerical-vector sequence modeling approaches such as Decision Transformers~\cite{chen2021decision} suffer from representation mismatch across heterogeneous modules and delayed credit assignment under scalar return-to-go conditioning. In contrast, grounding language models in executable action constraints~\cite{ahn2022saycan} and structured tool/API command generation~\cite{schick2023toolformer} has proven effective for adhering to formal protocols and suppressing out-of-distribution (OOD) execution errors.

Despite these advances, most existing LLM-based robotic systems adopt monolithic architectures tightly coupled to a specific foundation model family, fixed retrieval backend, or single training algorithm, hindering systematic progression from foundational representation verification to autonomous on-device edge deployment. Grounded in a Spatio-Temporal Dec-POMDP formulation~\cite{oliehoek2016concise}, our work establishes a pluggable functional block architecture and formal requirements specification (\texttt{REQ-1.1}--\texttt{REQ-3.4}) that strictly decouples the core encoding and reasoning modules validated in our v1 Proof-of-Concept from incremental extension modules (spatio-temporal context retrieval, uncertainty filtering, and offline policy adaptation for compact edge language models guided by deterministic verifiers) through implementation-agnostic input--output interfaces.

\section{Proposed Method}
\label{sec:method}

\subsection{Problem Formulation: Textual Scene Memory Maintenance as a Spatio-Temporal Dec-POMDP}
\label{sec:formulation}
Although the long-term Textual Scene Memory (TSM) maintenance problem addressed in this work is physically executed by a single mobile robot, strict partial observability caused by discarding raw images for memory efficiency and spatio-temporal fragmentation imposed by the context-length limits of inference models render it structurally equivalent to a cooperative problem among multiple spatio-temporally distributed memory-updating agents. Viewed through our continual learning analogy in SLAM---where irreversible scene-to-text translation updates symbolic scene parameters without experience replay---we formalize long-term map maintenance as a \emph{Spatio-Temporal Decentralized Partially Observable Markov Decision Process (Dec-POMDP)}~\cite{oliehoek2016concise} defined by the tuple $\mathcal{G} = \langle \mathcal{N}, \mathcal{S}, \mathcal{A}_{\text{mem}}, P, \Omega, O, R, \gamma \rangle$:

\begin{itemize}
    \item \textbf{Spatio-Temporal Decision-Making Agents ($\mathcal{N} = \{1, \dots, n\}$):} Represents local memory-updating agents associated with distinct topological regions (zones or landmark nodes $e_m \in \mathcal{M}_t$) or different patrol timestamps and camera viewpoints ($v_A, v_B, \dots$). Because agents cannot directly share raw images across time or viewpoints, they cooperatively maintain map consistency through the shared external scene memory $\mathcal{M}_t$.
    \item \textbf{Event-Driven Discrete Time Axis ($t \in \mathcal{T}_{\text{trig}}$):} To prevent wasted computation and redundant false updates in static environments, the discrete time step $t$ advances only when an observation trigger gate (\texttt{REQ-1.2}) detects a significant environmental event, rather than at every raw sensor frame.
    \item \textbf{Global State Space ($\mathcal{S}$):} Denotes the true underlying 3D physical and semantic state $s_t \in \mathcal{S}$ of the entire environment (including true object placements, open/closed or on/off states, and change persistence across all rooms). During execution, agents cannot directly observe $s_t$.
    \item \textbf{Private Observation Space and Linguistic Scene Parameterization ($\Omega, O$):} At each triggered step $t$, the agent acquires multi-view raw images $I_t$ conditioned on its current viewpoint $v_i$ (\texttt{REQ-1.1}) and immediately discards $I_t$ (\texttt{REQ-1.3}) after mapping it via hybrid textual encoding (\texttt{REQ-1.4}--\texttt{1.6}) into a textual observation $o_t = o_t^{\text{free}} \oplus o_t^{\text{struct}} \in \Omega^i$ (comprising a free-form description and a discrete state list including \texttt{unknown} for out-of-view objects). Rather than acting as a passive log, $o_t$ functions as \emph{explicit linguistic scene parameters} queried by a frozen foundation model; thus, the observation function $O$ must encode invariant sufficient statistics at write time before knowing which specific state-change queries will arise in the future.
    \item \textbf{Memory Operation Action Space ($\mathcal{A}_{\text{mem}}$):} Conditioned on the current observation $o_t$ and retrieved historical memory context $\mathcal{R}_t$ (\texttt{REQ-2.2}), each agent selects a discrete memory operation command $a_t \in \mathcal{A}_{\text{mem}}$ (\texttt{REQ-3.3}) according to its policy $\pi_\theta(a_t \mid o_t, \mathcal{R}_t)$, choosing among definitive updates (\texttt{COMMIT\_ADD} / \texttt{COMMIT\_UPDATE}), deletion (\texttt{COMMIT\_REMOVE}), transient deferral (\texttt{DEFER\_TRANSIENT}), and noise rejection (\texttt{REJECT\_NOISE})~\cite{ahn2022saycan}.
    \item \textbf{Environment Dynamics and Delayed Reward ($P, R, \gamma$):} The transition dynamics $P$ govern intrinsic environmental changes, viewpoint transitions induced by robot navigation, and external memory state updates ($\mathcal{M}_{t-1} \rightarrow \mathcal{M}_t$) resulting from $a_t$. While the reward $r_t = R(s_t, a_t)$ evaluates long-horizon logical consistency across the entire map, true step-wise persistence labels are unavailable during deployment, creating a delayed credit assignment problem analogous to replay-free continual learning without deployment supervision. Our specification addresses this via deterministic logical consistency invariants and occlusion gating (\texttt{REQ-3.2}, \texttt{REQ-3.4}) to prevent catastrophic forgetting.
\end{itemize}

\subsection{Taxonomy of Intrinsic Scene Changes and Qualitative Spatial Abstraction}
\label{sec:taxonomy}

\subsubsection{Boundary with Geometric Representations: Principle of Qualitative Spatial Abstraction}
\label{sec:qualitative_boundary}
Simply converting 3D point clouds or floating-point object coordinates $[x, y, z]$ from conventional geometric SLAM into ASCII text strings constitutes mere \emph{coordinate serialization}. Because raw numerical strings fluctuate with minor localization drift or viewpoint offsets, temporal string comparisons break down and fail to support semantic reasoning.

Therefore, our textual encoding specification adheres to the core architectural invariant of \emph{eliminating high-precision metric coordinate lists and abstracting spatial layouts into qualitative, intuitive descriptions that humans and language models can semantically reason about}. Under unknown future change queries, encoding unconstrained free-form text alone yields high-variance scene parameters that drift across viewpoints, whereas rigid slots alone lack descriptive coverage. As one illustrative example of a design guideline that realizes this boundary, spatial distances, areas, and scales can be quantized into approximate expressions within two significant digits (e.g., \texttt{space\_1.2\_tatami} for ``a space of about 1.2 tatami mats,'' \texttt{approx\_35\_ft} for ``approximately 35~ft ahead,'' or qualitative adjacency to static landmarks). By filtering out metric measurement noise and extracting only static-landmark-anchored topological relations and discrete states, the encoder guarantees high viewpoint invariance ($\phi_{\text{struct}}(I_t^{(v_A)}) \approx \phi_{\text{struct}}(I_t^{(v_B)})$) and stable semantic inference across distinct camera viewpoints $v_A$ and $v_B$ without geometric reconstruction or model weight updates.

\subsubsection{Systematic Taxonomy of Intrinsic Scene Changes}
\label{sec:change_taxonomy}
Under the qualitative spatial abstraction principle, once extrinsic observation noise---such as camera viewpoint motion, illumination changes, and temporary occlusions (targeted by \texttt{REJECT\_NOISE})---is decoupled and discarded, intrinsic scene changes occurring within the physical environment are categorized into six distinct classes, as summarized in Table~\ref{tab:change_taxonomy}:

\begin{enumerate}
    \item \textbf{Geometric / Spatial Changes:} Physical shifts in the relative position, orientation, or occupied region of an existing object (e.g., moving or rotating a chair). Anchored to static landmarks via qualitative descriptions (e.g., two-significant-digit approximate shifts), persistent relocations are distinguished from viewpoint noise and handled via \texttt{COMMIT\_UPDATE}.
    \item \textbf{Topological / Relational Changes:} Qualitative transitions in connectivity, adjacency, or support predicates between objects or spatial regions (e.g., \texttt{on(bottle, desk)} $\rightarrow$ \texttt{on(bottle, floor)}). As a core structural cue in coordinate-free textual memory, topological transitions require prompt reflection via \texttt{COMMIT\_UPDATE}.
    \item \textbf{Semantic / Object-Level Changes:} The addition of a new object (\texttt{COMMIT\_ADD}), permanent removal (\texttt{COMMIT\_REMOVE}), or replacement of an object in the scene. Mistaking temporary occlusion for physical disappearance and issuing \texttt{COMMIT\_REMOVE} is the primary trigger of catastrophic forgetting in replay-free linguistic memory.
    \item \textbf{State \& Attribute Changes:} Changes in mutable internal states or functional attributes while the object's spatial position and identity remain unchanged (e.g., door opening \texttt{closed} $\rightarrow$ \texttt{open}, or appliance power \texttt{OFF} $\rightarrow$ \texttt{ON}). These are processed by updating discrete attribute slots within the canonical state encoder (Section~\ref{sec:architecture}).
    \item \textbf{Kinematic / Transient Changes:} Short-lived, non-persistent variations caused by dynamic agents such as pedestrians or temporarily parked carts. Committing transient entities to persistent memory creates ``ghost objects'' upon revisit; hence, they are filtered out via \texttt{DEFER\_TRANSIENT}.
    \item \textbf{Scale \& Granularity Changes:} Variations in recognition granularity caused by observation distance or resolution differences (e.g., perceiving a pile of books collectively versus individually). Normalizing entities along a unified semantic granularity suppresses false-positive object additions via \texttt{REJECT\_NOISE}.
\end{enumerate}

\begin{table}[t]
\centering
\caption{Systematic Taxonomy of Intrinsic Scene Changes and Corresponding Memory Operations ($\mathcal{A}_{\text{mem}}$)}
\label{tab:change_taxonomy}
\resizebox{\columnwidth}{!}{%
\begin{tabular}{llll}
\toprule
\textbf{Intrinsic Category} & \textbf{Primary Target} & \textbf{Manifestation in Canonical Description} & \textbf{Memory Action ($\mathcal{A}_{\text{mem}}$)} \\
\midrule
1. Geometric & Position / orientation & Qualitative/coarse shift relative to static landmarks & \texttt{COMMIT\_UPDATE} / \texttt{DEFER\_TRANSIENT} \\
2. Topological & Inter-object relations & Relational predicate transition (\texttt{on}, \texttt{adjacent\_to}) & \texttt{COMMIT\_UPDATE} \\
3. Semantic & Addition / removal & Entity set change (new appearance / permanent removal) & \texttt{COMMIT\_ADD} / \texttt{COMMIT\_REMOVE} \\
4. State \& Attribute & Open/close, ON/OFF & Discrete state slot update within the same entity ID & \texttt{COMMIT\_UPDATE} \\
5. Kinematic & Pedestrians / carts & Transient variation lacking revisit persistence & \texttt{DEFER\_TRANSIENT} / \texttt{REJECT\_NOISE} \\
6. Scale & Aggregation granularity & Entity splitting/merging (absorbed via normalization) & \texttt{REJECT\_NOISE} \\
\bottomrule
\end{tabular}%
}
\end{table}

\subsection{System Overview and Formal Functional Requirements Specification}
\label{sec:architecture}
\begin{figure*}[t]
    \centering
    \includegraphics[width=\textwidth]{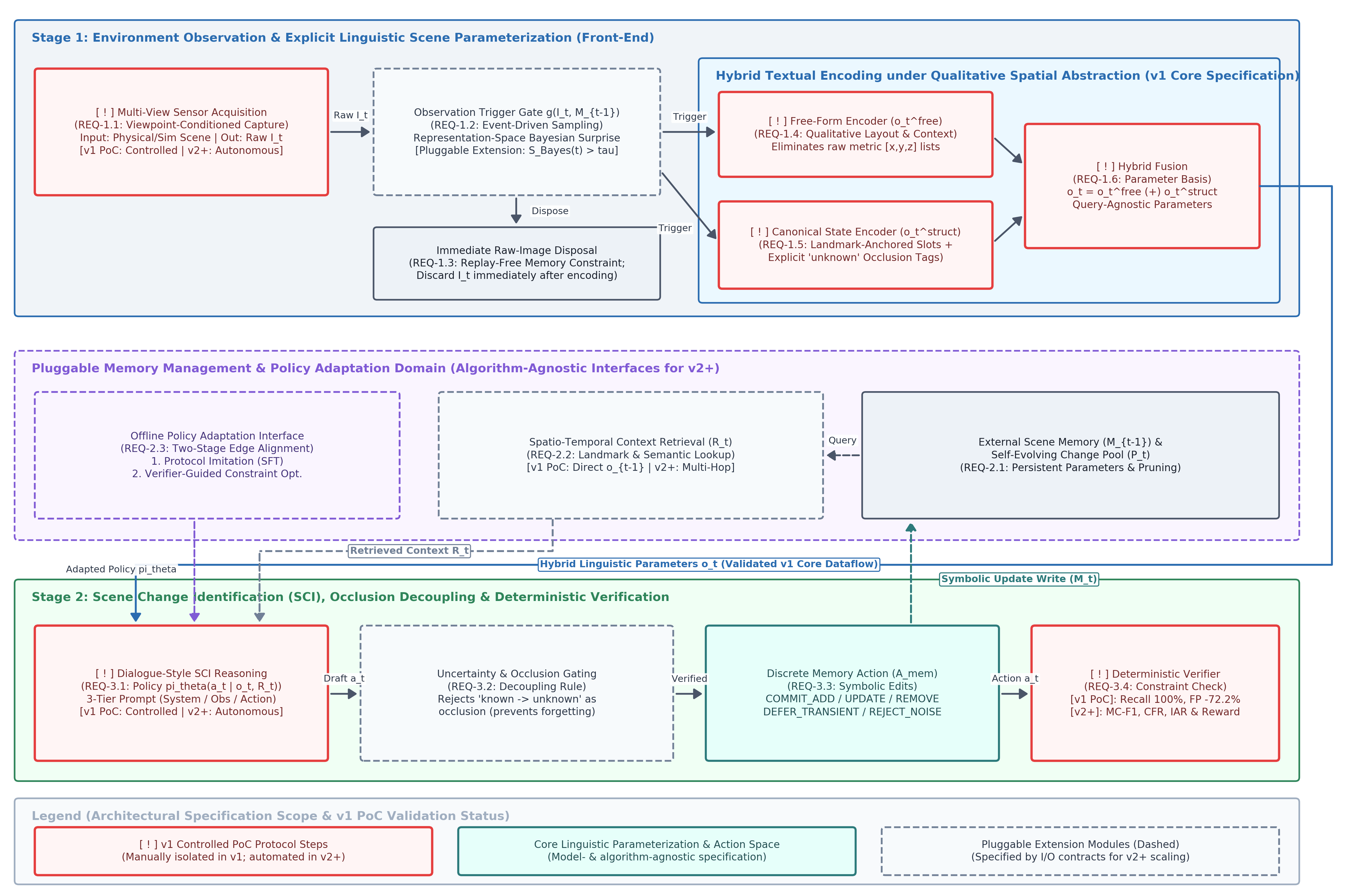}
    \caption{Overview of the modular Textual Scene Memory (TSM) requirements architecture. Solid-line blocks denote the core specification validated in v1, blocks marked with \texttt{[!]} indicate steps isolated under the controlled PoC protocol in Section~\ref{sec:experiments}, and dashed-line blocks represent pluggable extension interfaces (\texttt{REQ-1.2}, \texttt{REQ-2.1}--\texttt{2.3}, \texttt{REQ-3.2}) specified for long-horizon deployment in v2 and beyond.}
    \label{fig:pipeline}
\end{figure*}

Grounded in the Spatio-Temporal Dec-POMDP formulation and the qualitative spatial abstraction principle, Fig.~\ref{fig:pipeline} illustrates the overall architecture of \emph{Textual Scene Mapper (TSM)}, designed as a \emph{Pluggable Functional Architecture} composed of model- and algorithm-agnostic functional blocks. Rather than locking the system into a specific vision-language model, retrieval database, or policy optimization library, this specification defines strict input--output interface contracts and functional invariants (\texttt{REQ-1.1}--\texttt{REQ-3.4}) across three architectural domains, summarized in Table~\ref{tab:req_spec}.

\begin{table*}[t]
\centering
\caption{Summary of Functional Requirements Specification (\texttt{REQ-1.1}--\texttt{REQ-3.4}) and Input--Output Contracts}
\label{tab:req_spec}
\renewcommand{\arraystretch}{1.22}
\begin{tabularx}{\textwidth}{>{\raggedright\arraybackslash}p{0.10\textwidth} >{\raggedright\arraybackslash}p{0.23\textwidth} >{\raggedright\arraybackslash}p{0.23\textwidth} >{\raggedright\arraybackslash}X}
\toprule
\textbf{Req.\ ID} & \textbf{Functional Module} & \textbf{Input $\rightarrow$ Output Contract} & \textbf{Functional Invariant \& Theoretical Rationale} \\
\midrule
\texttt{REQ-1.1} & Multi-View Sensor Acquisition & Physical/Sim Scene $\rightarrow$ Viewpoint-tagged raw image $I_t^{(v_i)}$ & Captures local spatial observations across patrol viewpoints without requiring global metric map alignment. \\
\texttt{REQ-1.2} & Observation Trigger Gate (Ext.) & $(I_t, \mathcal{M}_{t-1}) \rightarrow \text{Trigger} \in \{0, 1\}$ & Suppresses redundant encoding in static scenes via representation-space surprise ($S_{\text{Bayes}}(t) > \tau_{\text{Bayes}}$). \\
\texttt{REQ-1.3} & Immediate Raw-Image Disposal & $I_t \rightarrow \emptyset$ (post-encoding) & Enforces strict replay-free memory footprint; prohibits storing raw pixels or dense visual embeddings. \\
\texttt{REQ-1.4} & Free-Form Encoder ($o_t^{\text{free}}$) & $I_t \rightarrow o_t^{\text{free}}$ (qualitative text) & Verbalizes open-ended layout/context under qualitative abstraction; eliminates raw floating-point $[x,y,z]$ lists. \\
\texttt{REQ-1.5} & Canonical State Encoder ($o_t^{\text{struct}}$) & $I_t \rightarrow \{(c_k, \ell_k, \sigma_{k,t})\}$ & Grounds discrete state slots in static landmarks and emits explicit \texttt{unknown} tags for occluded/out-of-view items. \\
\texttt{REQ-1.6} & Hybrid Parameter Fusion & $(o_t^{\text{free}}, o_t^{\text{struct}}) \rightarrow o_t = o_t^{\text{free}} \oplus o_t^{\text{struct}}$ & Constructs query-agnostic explicit linguistic parameters balancing descriptive coverage and viewpoint invariance. \\
\midrule
\texttt{REQ-2.1} & External Scene Memory \& Pool (Ext.) & $(\mathcal{M}_{t-1}, a_t) \rightarrow (\mathcal{M}_t, \mathcal{P}_t)$ & Persists landmark-anchored linguistic parameters and prunes noisy change records via revisit utility feedback. \\
\texttt{REQ-2.2} & Spatio-Temporal Retrieval (Ext.) & $(o_t, \mathcal{M}_{t-1}) \rightarrow \mathcal{R}_t$ & Retrieves landmark-matched historical context within foundation model context limits (degenerates to $o_{t-1}$ in 2-step PoC). \\
\texttt{REQ-2.3} & Offline Policy Adaptation (Ext.) & $\mathcal{D}_{\text{offline}} \rightarrow \pi_\theta$ & Adapts compact on-device language models via protocol imitation and verifier-guided optimization without online risk. \\
\midrule
\texttt{REQ-3.1} & Dialogue-Style SCI Reasoning & $(o_t, \mathcal{R}_t) \rightarrow \text{Draft } a_t$ & Infers intrinsic state transitions in a shared semantic space via a 3-tier prompt (System / Observation / Action). \\
\texttt{REQ-3.2} & Uncertainty \& Occlusion Gating (Ext.) & $(\text{Draft } a_t, o_t, \mathcal{R}_t) \rightarrow \text{Verified } a_t$ & Decouples \texttt{known} $\rightarrow$ \texttt{unknown} transitions as observation occlusion rather than removal, preventing catastrophic forgetting. \\
\texttt{REQ-3.3} & Discrete Memory Action Controller & $\text{Verified } a_t \rightarrow a_t \in \mathcal{A}_{\text{mem}}$ & Emits symbolic parameter edits (\texttt{COMMIT\_ADD/UPDATE/REMOVE}, \texttt{DEFER\_TRANSIENT}, \texttt{REJECT\_NOISE}). \\
\texttt{REQ-3.4} & Deterministic Verifier & $(u_t, a_t, \mathcal{M}_{t-1}) \rightarrow r \in \{-1, 0, +1\}$ & Enforces schema validity, ID grounding, and \texttt{unknown}-deletion prohibition; computes $\text{MC-F1}$, $\text{CFR}$, and $\text{IAR}$. \\
\bottomrule
\end{tabularx}
\end{table*}

The pipeline operates across three functional domains:
\begin{enumerate}
    \item \textbf{Stage 1: Environment Observation and Explicit Linguistic Parameterization (Front-End, \texttt{REQ-1.1}--\texttt{REQ-1.6}):} Multi-view raw images $I_t$ acquired by the sensor module (\texttt{REQ-1.1}) are admitted to encoding only when the event-driven observation trigger gate (\texttt{REQ-1.2}) detects a candidate change, and are immediately discarded upon encoding (\texttt{REQ-1.3}) to guarantee a replay-free memory footprint. Under the qualitative spatial abstraction principle (Section~\ref{sec:qualitative_boundary}), the front-end drives two parallel encoders without knowing future state-change queries: a \emph{free-form description encoder} ($o_t^{\text{free}}$, \texttt{REQ-1.4}) that flexibly verbalizes spatial layouts and visual attributes without raw coordinate lists, and a \emph{canonical state encoder} ($o_t^{\text{struct}} = \{(c_k, \ell_k, \sigma_{k,t})\}$, \texttt{REQ-1.5}) that extracts qualitative spatial relations $\ell_k$ anchored to static landmarks, discrete object states, and explicit occlusion tags (\texttt{unknown}) based on a canonical vocabulary schema. Their outputs are fused into a unified hybrid observation $o_t = o_t^{\text{free}} \oplus o_t^{\text{struct}}$ (\texttt{REQ-1.6}), forming an explicit linguistic scene parameter basis.
    \item \textbf{Pluggable Memory Management and Policy Adaptation Domain (\texttt{REQ-2.1}--\texttt{REQ-2.3}):} Viewed through our continual-learning-in-SLAM analogy, this back-end domain governs long-horizon parameter persistence and edge policy refinement. It comprises an \emph{external scene memory storage} ($\mathcal{M}_{t-1}$) paired with a \emph{self-evolving change memory pool} ($\mathcal{P}_t$) that retains landmark-anchored parameters and prunes noisy transition examples via revisit utility feedback (\texttt{REQ-2.1}); a \emph{spatio-temporal context retrieval} module ($\mathcal{R}_t$) that fetches relevant historical memories by landmark proximity and semantic similarity (\texttt{REQ-2.2}); and an \emph{offline policy adaptation} interface (\texttt{REQ-2.3}) that distills protocol compliance into compact on-device language models. By enforcing strict input--output contracts, this domain supports degenerate operation in two-step change identification tasks by passing the prior observation $o_{t-1}$ directly as $\mathcal{R}_t$.
    \item \textbf{Stage 2: Scene Change Identification (SCI), Occlusion Decoupling, and Verification (\texttt{REQ-3.1}--\texttt{REQ-3.4}):} The \emph{dialogue-style SCI reasoning} module (\texttt{REQ-3.1}) compares $\mathcal{R}_t$ against $o_t$ to propose candidate memory edits. Next, the \emph{uncertainty and occlusion gating} module (\texttt{REQ-3.2}) enforces a structural decoupling invariant: transitions from a known state to an unobserved state (\texttt{known} $\rightarrow$ \texttt{unknown}) are classified as extrinsic viewpoint or foreground occlusions rather than physical removals, directly preventing catastrophic forgetting of accumulated symbolic parameters. Finally, the \emph{discrete memory action controller} (\texttt{REQ-3.3}) outputs verified symbolic commands $a_t \in \mathcal{A}_{\text{mem}}$, and the \emph{deterministic verifier} (\texttt{REQ-3.4}) audits schema compliance, entity-ID grounding, and safety constraints against ground truth.
\end{enumerate}

\subsection{Dialogue-Style State-Change Reasoning Protocol and Phased Validation Scope}
\label{sec:dialogue_protocol}
Within \texttt{REQ-3.1}, the memory-updating policy $\pi_\theta(a_t \mid o_t, \mathcal{R}_t)$ of the Spatio-Temporal Dec-POMDP is realized via structured natural language interaction (\emph{Dialogue-Style Sequence Modeling})~\cite{schick2023toolformer}:
\begin{itemize}
    \item \textbf{Three-Tier Dialogue Prompt Structure:} The inference model is conditioned on a three-tier prompt contract comprising: (1) a \emph{System Prompt} defining the agent's role, canonical schema rules, and the hard safety invariant prohibiting the classification of \texttt{unknown} states as physical disappearance (\texttt{COMMIT\_REMOVE}); (2) an \emph{Observation Prompt (User Turn)} presenting the current hybrid parameters $o_t$ alongside the retrieved historical context $\mathcal{R}_t$; and (3) an \emph{Action Prompt (Assistant Turn)} eliciting step-by-step logical chain-of-thought reasoning~\cite{wei2022chainofthought} followed by discrete memory operation commands in $\mathcal{A}_{\text{mem}}$.
    \item \textbf{Separation Between v1 PoC Validation and v2+ Autonomous Extensions:} In Fig.~\ref{fig:pipeline}, solid-line blocks define the core linguistic parameterization and reasoning specification. Among these, the five steps marked with \texttt{[!]} (\texttt{REQ-1.1}, \texttt{REQ-1.4}, \texttt{REQ-1.5}, \texttt{REQ-1.6}, and \texttt{REQ-3.1}/\texttt{3.4}) are executed under a strictly controlled manual PoC protocol in Section~\ref{sec:experiments} (Sets~1--10) to isolate representation invariance from engineering noise. Conversely, the dashed-line extension blocks (\texttt{REQ-1.2}, \texttt{REQ-2.1}--\texttt{2.3}, \texttt{REQ-3.2}) and programmatic automation of the \texttt{[!]} steps represent incremental engineering phases for v2 and beyond under identical input--output contracts, whose formal mathematical specifications and theoretical rationale are detailed below.
\end{itemize}

\subsection{Formal Mathematical Specifications and Theoretical Rationale for Pluggable Modules}
\label{sec:extension_specs}

\subsubsection{Event-Driven Trigger Gate (\texttt{REQ-1.2}) and Self-Evolving Change Memory Pool (\texttt{REQ-2.1}, \texttt{REQ-2.2})}
\label{sec:pool_trigger_spec}
To maintain long-horizon consistency without retaining raw images, the pluggable observation gating and memory management modules satisfy the following formal specifications:
\begin{itemize}
    \item \textbf{Candidate Event-Driven Gate via Representation-Space Bayesian Surprise (\texttt{REQ-1.2}):} To avoid frame-by-frame VLM captioning and discrete probability saturation, raw observations $I_t$ are projected onto a similarity weight vector $\mathbf{w}_t \in \Delta^{K-1}$ over coordinate-free deterministic typological descriptors (e.g., Bag-of-Words) to smooth domain noise. Defining Bayesian surprise in this representation space as the Kullback--Leibler (KL) divergence between the sequential prior $P(\theta \mid \mathbf{w}_{<t})$ predicted from past trajectories and the posterior $P(\theta \mid \mathbf{w}_{\le t})$ after observing $\mathbf{w}_t$:
    \begin{equation}
        S_{\text{Bayes}}(t) = D_{\text{KL}}\Bigl( P(\theta \mid \mathbf{w}_{\le t}) \;\Big\Vert\; P(\theta \mid \mathbf{w}_{<t}) \Bigr),
    \end{equation}
    downstream hybrid textual encoding is triggered only when $S_{\text{Bayes}}(t) > \tau_{\text{Bayes}}$, balancing computational efficiency with selective extraction of semantically meaningful events.
    \item \textbf{Self-Evolving Change Memory Pool and Erroneous Transition Pruning (\texttt{REQ-2.1}, \texttt{REQ-2.2}):} To prevent the accumulation of erroneous change records caused by perceptual misclassifications or viewpoint noise, a storage structure termed the \emph{Self-Evolving Change Memory Pool} $\mathcal{P}_t = \{\kappa_1, \dots, \kappa_{|\mathcal{P}_t|}\}$ with capacity limit $C_{\text{pool}}$ is embedded within the external scene memory $\mathcal{M}_t$. Each entry $\kappa_i = (\text{id}_i, \delta_i, \text{ctx}_i, \text{verdict}_i, U_i, n_i, t_{\text{last}, i})$ stores the canonical state transition $\delta_i$, static-landmark-anchored qualitative spatial context $\text{ctx}_i$, persistence label $\text{verdict}_i \in \{\texttt{Persistent}, \texttt{Transient}, \texttt{Occlusion}\}$, utility score $U_i \in [0, 1]$ reflecting historical contribution to map consistency, and cumulative reference count $n_i$. When retrieving past change examples for $\mathcal{R}_t$ (\texttt{REQ-2.2}), entries are ranked via a composite priority score integrating semantic similarity, utility, frequency penalty, and a novelty bonus $B_{\text{nov}}(\kappa_i)$:
    \begin{equation}
        \text{Score}(q_t, \kappa_i) = w_{\text{sim}} \cdot \text{Sim}(q_t, \kappa_i) + w_{\text{util}} \cdot U_i - w_{\text{freq}} \cdot \ln(1 + n_i) + w_{\text{nov}} \cdot B_{\text{nov}}(\kappa_i).
    \end{equation}
    Combined with hierarchical cluster-diverse selection across object categories to prevent retrieval bias, post-hoc revisit verification feedback ($\text{Credit}_t \in \{0, +1\}$) updates utility scores via $U_i \leftarrow (1 - \eta)U_i + \eta \cdot \text{Credit}_t$, naturally pruning noisy transition prompts over time while retaining reliable transition knowledge.
\end{itemize}

\subsubsection{Two-Stage Offline Policy Adaptation (\texttt{REQ-2.3}) and Deterministic Verifier (\texttt{REQ-3.4})}
\label{sec:adaptation}
To support autonomous real-time operation without cloud latency or bandwidth overhead in v2 and beyond, \texttt{REQ-2.3} specifies an offline policy adaptation interface tailored to compact on-device Small Language Models (SLMs, 2B--7B parameters)~\cite{bai2023qwen, grattafiori2024llama3}. Rather than relying on risky online trial-and-error in physical environments, the specification defines a two-stage offline alignment contract:
\begin{enumerate}
    \item \textbf{Stage 1: Return-Agnostic Autoregressive SFT (Protocol Distillation):} On an offline dialogue trajectory dataset $\mathcal{D}_{\text{SFT}} = \{\tau^{(k)}\}_{k=1}^{M_{\text{SFT}}}$, the model minimizes the negative log-likelihood of imitating ground-truth memory operation responses $a_t^{(k), i} \in \mathcal{A}_{\text{mem}}$ conditioned on the dialogue history $\tau_{\le (u_t^{(k), i})}^{(k)}$:
    \begin{equation}
        \mathcal{L}_{\text{SFT}}(\theta) = -\frac{1}{M_{\text{SFT}}} \sum_{k=1}^{M_{\text{SFT}}} \sum_{t=1}^{T^{(k)}} \sum_{i=1}^{N} \log P_\theta \left( a_t^{(k), i} \;\middle|\; \tau_{\le (u_t^{(k), i})}^{(k)} \right).
    \end{equation}
    \item \textbf{Stage 2: OOD Boundary Filtering and GRPO Alignment:} Constructing a boundary subset $\mathcal{D}_{\text{OOD}}$ containing hard cases where the SFT policy $\pi_{\theta_{\text{SFT}}}$ commits misclassifications or constraint violations, the policy maximizes the Group Relative Policy Optimization (GRPO)~\cite{shao2024deepseekmath} objective over $G$ sampled candidate responses $\{a_j^i\}_{j=1}^G$ using group-relative advantages $\hat{A}_{j,s}$:
    \begin{multline}
        \mathcal{J}_{\text{GRPO}}(\theta) = \mathbb{E}_{u_t^i \sim \mathcal{D}_{\text{OOD}}, \, \{a_j^i\}_{j=1}^G \sim \pi_{\theta_{\text{SFT}}}} \Biggl\{ \frac{1}{G} \sum_{j=1}^{G} \frac{1}{|a_j^i|} \sum_{s=1}^{|a_j^i|} \min \Bigl[ \rho_{j,s}(\theta) \hat{A}_{j,s}, \\
        \operatorname{clip}\bigl(\rho_{j,s}(\theta), 1-\epsilon, 1+\epsilon\bigr) \hat{A}_{j,s} \Bigr] - \beta \mathbb{D}_{\text{KL}}\bigl[\pi_\theta \,\Vert\, \pi_{\text{ref}}\bigr] \Biggr\}.
    \end{multline}
\end{enumerate}

The \emph{Deterministic Rule-Based Verifier} (\texttt{REQ-3.4}) responsible for both GRPO reward assignment and automated benchmark evaluation enforces the valid action constraint set $\mathcal{A}_{\text{valid}}(u_t^i, \mathcal{M}_{t-1}) \subseteq \mathcal{A}_{\text{mem}}$~\cite{ahn2022saycan}---requiring (i) valid JSON schema~\cite{jsonschema2020} compliance, (ii) reference only to existing entity IDs in $\mathcal{M}_{t-1}$ or $o_t$, and (iii) strict prohibition of issuing \texttt{COMMIT\_REMOVE} for entities in the \texttt{unknown} state---and computes a three-tier scalar reward against ground truth $\hat{a}_t^i$:
\begin{equation}
    r(u_t^i, a_t^i) = \begin{cases}
        +1, & \text{if } a_t^i = \hat{a}_t^i \text{ and } a_t^i \in \mathcal{A}_{\text{valid}}, \\
        0, & \text{if } a_t^i \neq \hat{a}_t^i \text{ and } a_t^i \in \mathcal{A}_{\text{valid}}, \\
        -1, & \text{if } a_t^i \notin \mathcal{A}_{\text{valid}},
    \end{cases}
\end{equation}
where $+1$ rewards exact ground-truth compliance, $0$ grants safe deferral without catastrophic forgetting, and $-1$ penalizes syntactic errors, hallucinated entity IDs, or erroneous deletion of \texttt{unknown} objects. Furthermore, during inference, if greedy decoding produces an invalid operation ($a_t^i \notin \mathcal{A}_{\text{valid}}$), fallback resampling from a $\text{top-}p / \text{top-}k$ truncated distribution prevents logical corruption of the scene memory.

\subsubsection{Information-Theoretic Rationale and Long-Horizon Evaluation Metrics}
\label{sec:metrics_spec}
The theoretical justification for why our architecture achieves accurate spatial reasoning while discarding raw images lies in decoupling the end-to-end information transformation into two distinct stages:
\begin{itemize}
    \item \textbf{Lossy Domain-Invariant Compression via Linguistic Scene Parameterization (Front-End):} Mapping visual observations $I_t$ to hybrid textual observations $o_t = o_t^{\text{free}} \oplus o_t^{\text{struct}}$ ($H(o_t) \ll H(I_t)$) constitutes intentional lossy compression~\cite{cover2006elements} that identifies explicit linguistic scene parameters prior to knowing future state-change queries. By eliminating raw coordinate serialization in favor of qualitative spatial descriptions (e.g., two-significant-digit approximations) and static-landmark-anchored discrete state schemas (including \texttt{unknown} tags), the front-end acts as a structured information bottleneck that selectively discards extrinsic observation noise (metric drift, illumination shifts, and viewpoint offsets) while preserving query-agnostic sufficient statistics despite the absence of experience replay.
    \item \textbf{Lossless Shared Semantic Space Embedding and Discrete Stability--Plasticity Control (Back-End):} The dialogue prompt $u_t$ integrating canonical scene parameters and retrieved historical context is tokenized into a lossless discrete token sequence and embedded into a unified pretrained semantic space $\mathbb{R}^d$~\cite{vaswani2017attention}. Unlike numerical-vector sequence modeling~\cite{chen2021decision}, which suffers from representation mismatch across heterogeneous modules, operating within a shared linguistic parameter space allows a frozen foundation model to leverage pretrained commonsense reasoning without geometric reconstruction or model weight updates. Furthermore, this architecture directly resolves the four continual-learning challenges identified in Section~\ref{sec:rw_forgetting}: static-landmark anchoring and canonical slots regularize discrete parameter edits against lexical drift (preserving entity identity without continuous weight regularization), while structured uncertainty gating (\texttt{known} $\rightarrow$ \texttt{unknown} $\mapsto$ \texttt{REJECT\_NOISE} / \texttt{DEFER\_TRANSIENT}) decouples observation shifts from true environmental changes and prevents catastrophic forgetting even without ground-truth supervision at deployment.
\end{itemize}

To quantitatively evaluate long-horizon performance once the fully automated pipeline (programmatic execution of the \texttt{[!]} steps and dashed extension blocks in Fig.~\ref{fig:pipeline}) is deployed in v2 and beyond, we define three standard evaluation metrics:
\begin{enumerate}
    \item \textbf{Map Consistency F1 Score ($\text{MC-F1}$):} The harmonic mean of precision and recall between the final scene memory $\mathcal{M}_T$ at the end of a long-horizon patrol and the true 3D semantic state set $\mathcal{S}_T^*$ reflecting only persistent changes ($\text{MC-F1} = 2 \cdot \text{Precision} \cdot \text{Recall} / (\text{Precision} + \text{Recall})$).
    \item \textbf{Catastrophic Forgetting Rate ($\text{CFR}$):} The proportion of known objects that still physically exist in the environment but became \texttt{unknown} due to viewpoint shifts or occlusions and were erroneously deleted (\texttt{COMMIT\_REMOVE}) or overwritten (lower is better).
    \item \textbf{Invalid / Out-of-Distribution Action Rate ($\text{IAR}$):} The fraction of generated memory operation commands $a_t$ that violate the deterministic verifier's valid action set $\mathcal{A}_{\text{valid}}(u_t, \mathcal{M}_{t-1})$, defined as $\text{IAR} = \frac{1}{T} \sum_{t=1}^T \mathbb{I}(a_t \notin \mathcal{A}_{\text{valid}}(u_t, \mathcal{M}_{t-1}))$.
\end{enumerate}

\subsubsection{Phased Migration Design Toward Full Programmatic Automation}
\label{sec:automation_roadmap}
Building upon the empirically validated effectiveness of hybrid textual encoding, the manual \texttt{[!]} steps and dashed extension blocks in Fig.~\ref{fig:pipeline} are scheduled for complete programmatic automation in v2 and beyond through four modular engineering replacements that preserve the exact input--output contracts specified in Table~\ref{tab:req_spec}, as summarized in Table~\ref{tab:automation_roadmap}:

\begin{table*}[t]
\centering
\caption{Migration Design from Controlled Manual Protocol (v1) to Fully Automated Pipeline (v2 and Beyond)}
\label{tab:automation_roadmap}
\renewcommand{\arraystretch}{1.25}
\begin{tabularx}{\textwidth}{>{\raggedright\arraybackslash}p{0.21\textwidth} >{\raggedright\arraybackslash}p{0.27\textwidth} >{\raggedright\arraybackslash}X}
\toprule
\textbf{Pipeline Stage (Fig.~\ref{fig:pipeline})} & \textbf{Controlled Protocol in This Study (v1)} & \textbf{Programmatic Implementation Design (v2 and Beyond)} \\
\midrule
1. Multi-View Observation (\texttt{REQ-1.1}, \texttt{1.2}) & Manual camera placement and state rendering in Isaac Sim GUI & \textbf{Headless Automated Data Generation \& Surprise Gating:} Automated camera trajectory sampling and randomized object state transitions with ground-truth JSON export, paired with Bayesian surprise event gating ($S_{\text{Bayes}}(t) > \tau_{\text{Bayes}}$). \\
2. Textual Encoding (\textbf{P1}, \textbf{P2}) \& Fusion (\texttt{REQ-1.4}--\texttt{1.6}) & Manual image attachment in chat UI and manual concatenation of $o_t^{\text{free}} \oplus o_t^{\text{struct}}$ & \textbf{Asynchronous Batch Inference \& Schema-Constrained Decoding:} Parallel VLM batch encoding with JSON Schema constrained decoding and automated vocabulary expansion from external state datasets (OSCaR~\cite{nguyen2024oscar}, ChangeIt-Frames~\cite{soucek2022look}). \\
3. SCI Reasoning (\textbf{P3}) \& Memory (\texttt{REQ-2.1}--\texttt{2.3}, \texttt{3.1}--\texttt{3.3}) & Manual copy-pasting of two-step notes into chat UI for direct text comparison & \textbf{Autonomous Reasoning Loop \& Spatio-Temporal Retrieval:} Automated read/write control over external memory $\mathcal{M}_t$ and memory pool $\mathcal{P}_t$, uncertainty filtering rejecting \texttt{known} $\rightarrow$ \texttt{unknown} transitions, and two-stage offline edge SLM adaptation. \\
4. Quantitative Evaluation (\texttt{REQ-3.4}) & Experimenter audit and tallying under deterministic counting rules & \textbf{Deterministic Verifier Evaluation Harness:} Automated matching between structured outputs and simulation ground-truth JSON to continuously compute $\text{MC-F1}$, $\text{CFR}$, and $\text{IAR}$. \\
\bottomrule
\end{tabularx}
\end{table*}

\section{Foundational Proof-of-Concept Validation of Core Requirements}
\label{sec:experiments}

In this section, we empirically validate the foundational core of our architectural specification---namely, hybrid textual encoding ($o_t = o_t^{\text{free}} \oplus o_t^{\text{struct}}$, \texttt{REQ-1.4}--\texttt{1.6}) under strict raw-image disposal (\texttt{REQ-1.3}) and dialogue-style state-change reasoning (\texttt{REQ-3.1}, \texttt{REQ-3.4})---on a multi-view benchmark dataset in a 3D physics simulator. We detail the experimental setup, compared representations, deterministic counting rules, quantitative and qualitative findings, and the methodological justification explaining why our human-controlled evaluation protocol establishes rigorous validity as a Proof-of-Concept (PoC).

\subsection{Experimental Environment and Controlled Evaluation Protocol (Sets 1--10)}
\label{sec:exp_setup}

\subsubsection{Simulation Environment and Multi-View State-Change Dataset}
\label{sec:dataset}
\begin{figure*}[t]
    \centering
    \begin{minipage}[t]{0.485\textwidth}
        \centering
        \textbf{\small Kitchen Environment (Sets~1--5)}\\[3pt]
        \includegraphics[width=\textwidth]{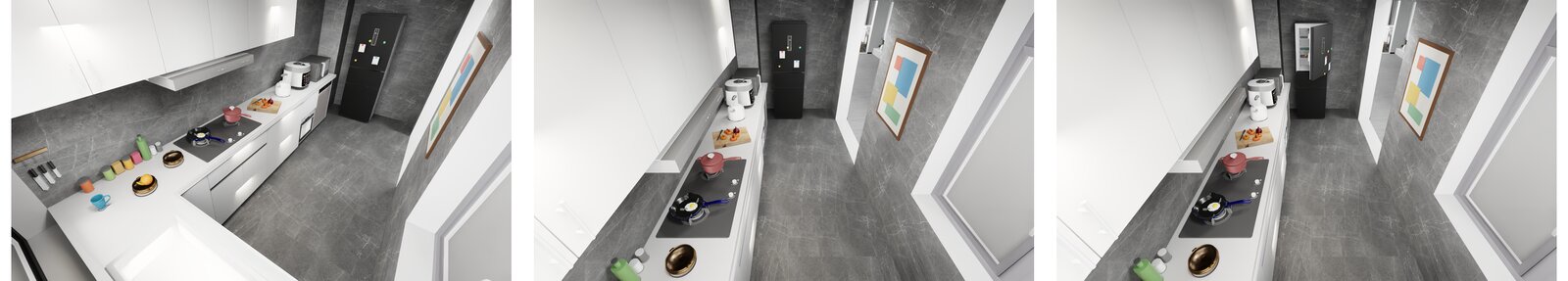}\\[1pt]
        {\scriptsize (a) Set~1: Refrigerator door (\texttt{closed} $\rightarrow$ \texttt{open})}\\[3pt]
        \includegraphics[width=\textwidth]{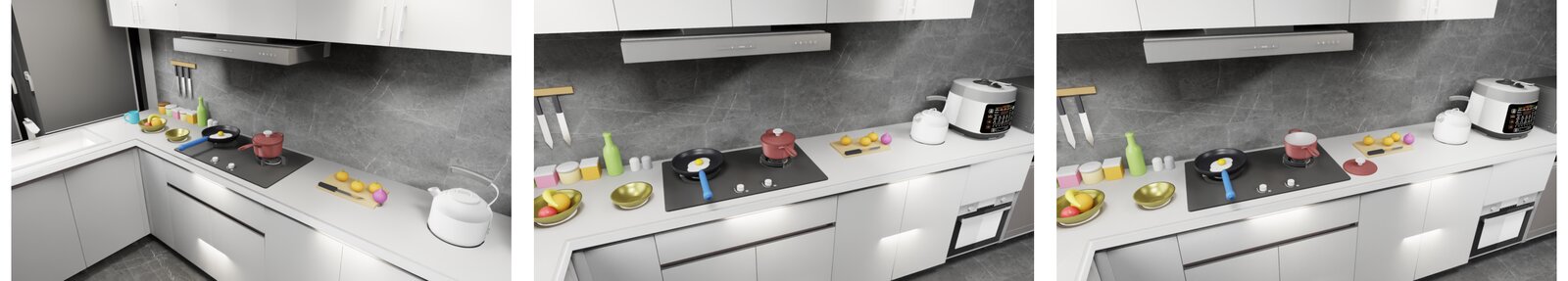}\\[1pt]
        {\scriptsize (b) Set~2: Pot lid (\texttt{lid on} $\rightarrow$ \texttt{lid off})}\\[3pt]
        \includegraphics[width=\textwidth]{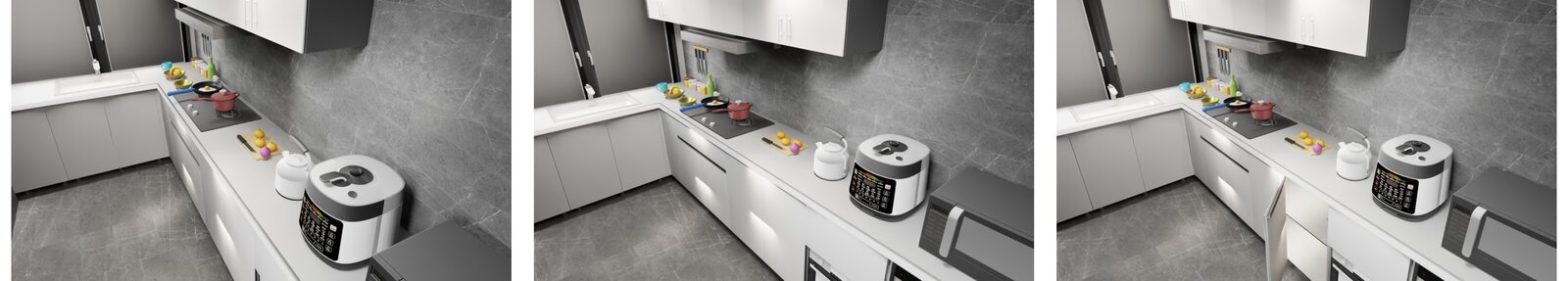}\\[1pt]
        {\scriptsize (c) Set~3: Cabinet door (\texttt{closed} $\rightarrow$ \texttt{open})}\\[3pt]
        \includegraphics[width=\textwidth]{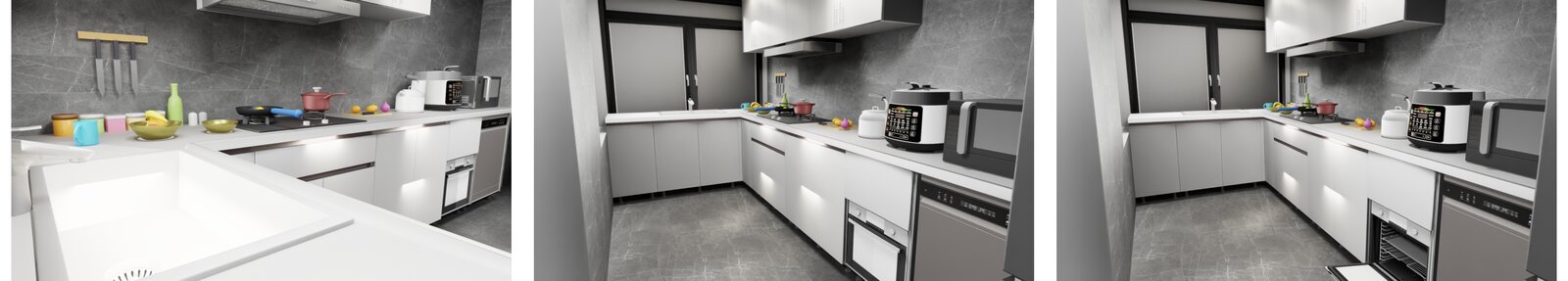}\\[1pt]
        {\scriptsize (d) Set~4: Oven door (\texttt{closed} $\rightarrow$ \texttt{open})}\\[3pt]
        \includegraphics[width=\textwidth]{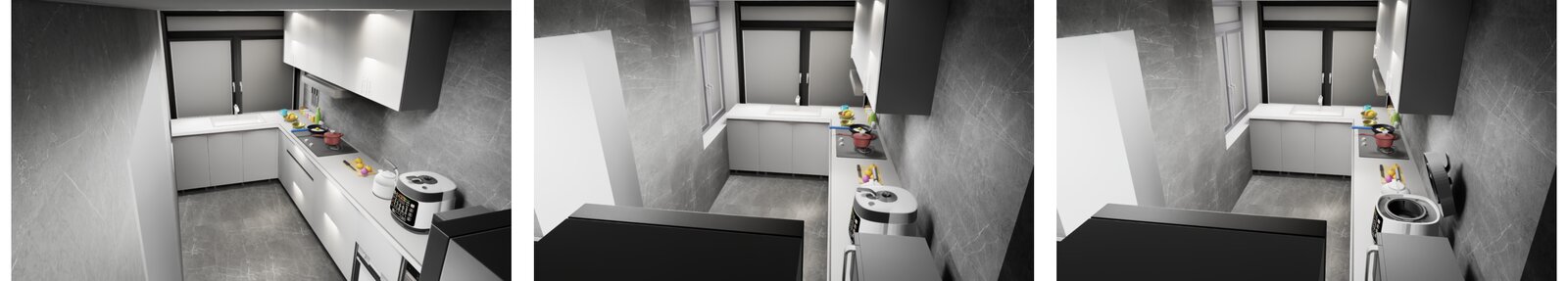}\\[1pt]
        {\scriptsize (e) Set~5: Rice cooker lid (\texttt{lid closed} $\rightarrow$ \texttt{lid open})}
    \end{minipage}
    \hfill
    \begin{minipage}[t]{0.485\textwidth}
        \centering
        \textbf{\small Bedroom Environment (Sets~6--10)}\\[3pt]
        \includegraphics[width=\textwidth]{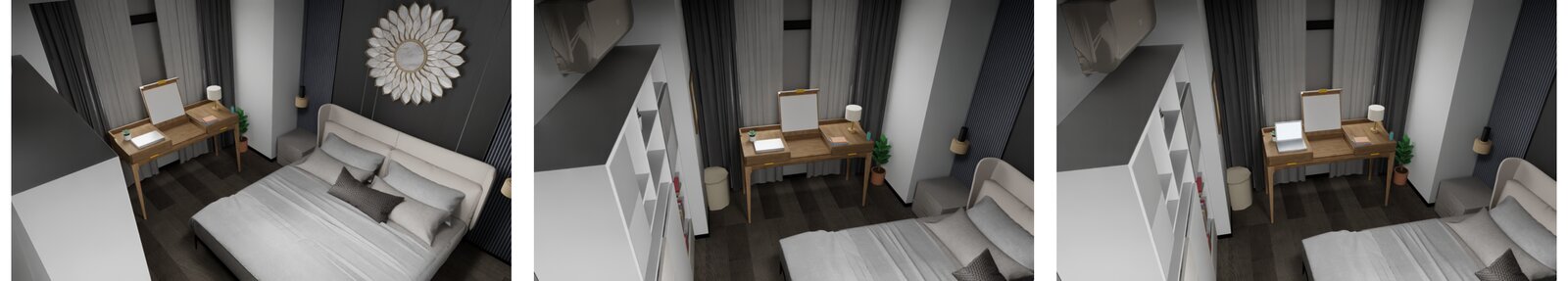}\\[1pt]
        {\scriptsize (f) Set~6: Laptop lid (\texttt{closed} $\rightarrow$ \texttt{open})}\\[3pt]
        \includegraphics[width=\textwidth]{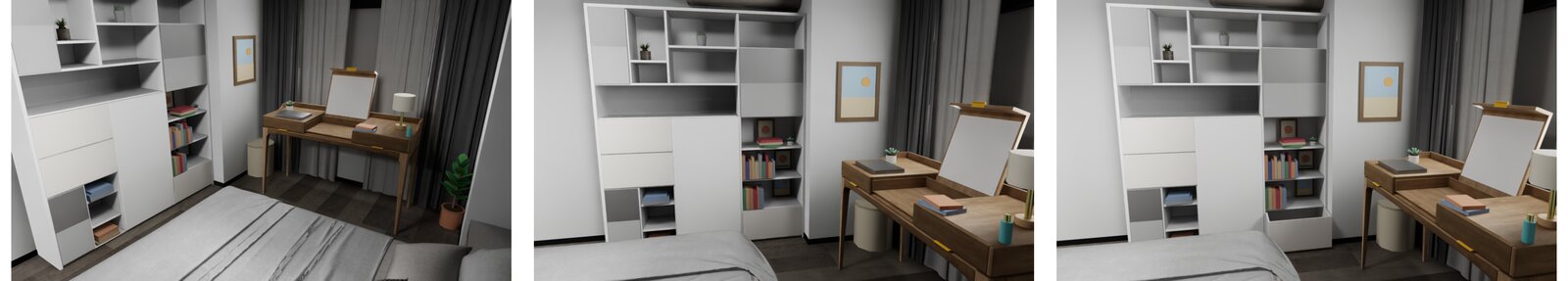}\\[1pt]
        {\scriptsize (g) Set~7: Cupboard drawer (\texttt{closed} $\rightarrow$ \texttt{open})}\\[3pt]
        \includegraphics[width=\textwidth]{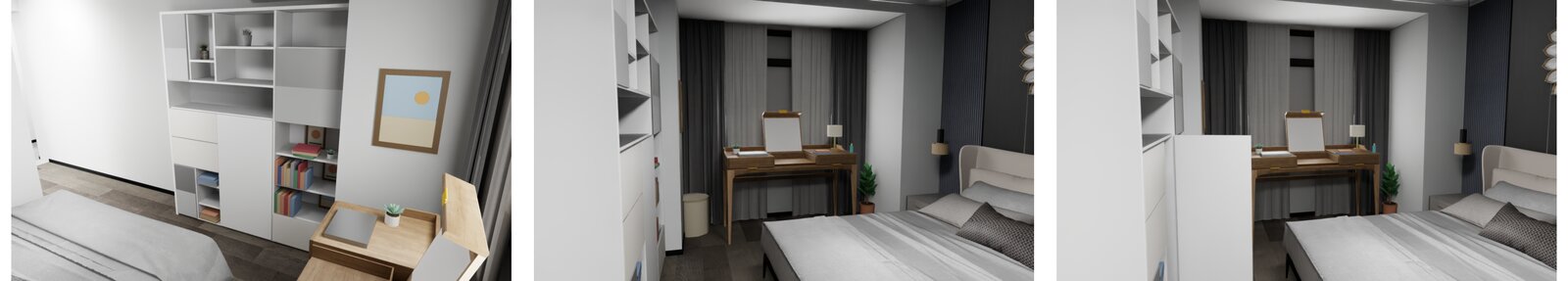}\\[1pt]
        {\scriptsize (h) Set~8: Cupboard door (\texttt{closed} $\rightarrow$ \texttt{open})}\\[3pt]
        \includegraphics[width=\textwidth]{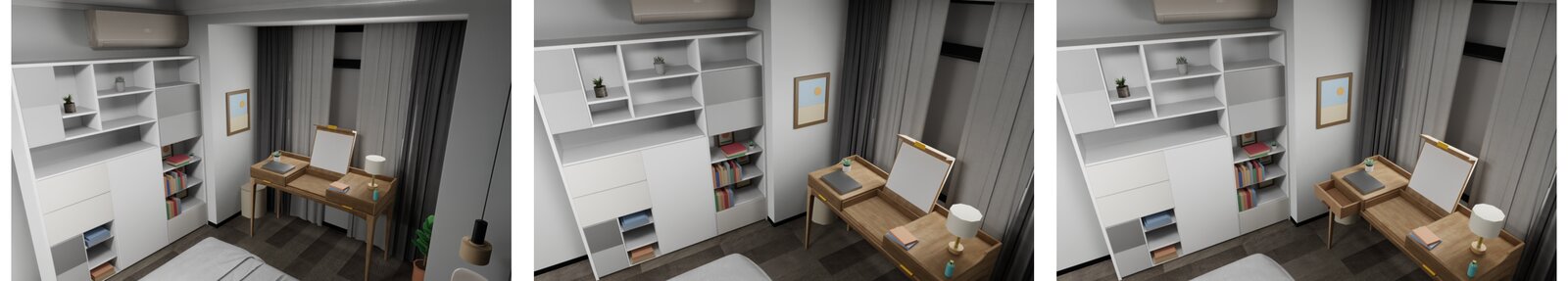}\\[1pt]
        {\scriptsize (i) Set~9: Desk drawer (\texttt{closed} $\rightarrow$ \texttt{open})}\\[3pt]
        \includegraphics[width=\textwidth]{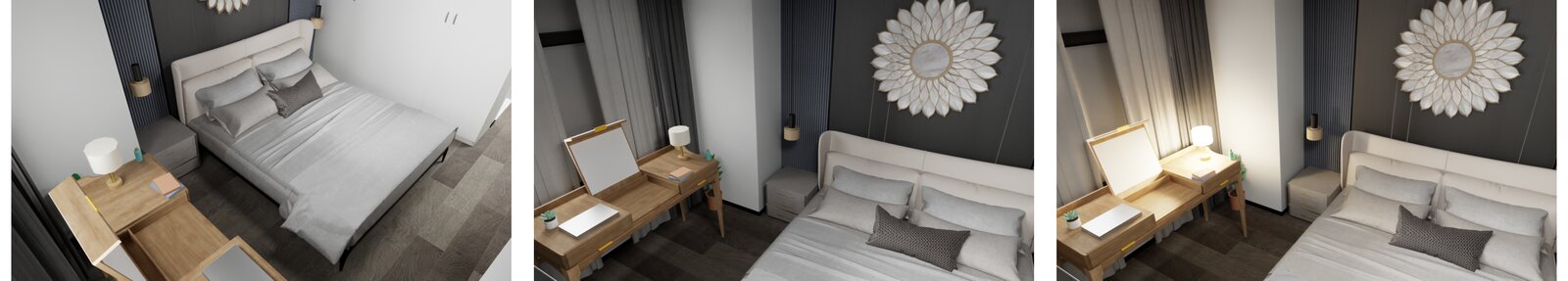}\\[1pt]
        {\scriptsize (j) Set~10: Table lamp (\texttt{off} $\rightarrow$ \texttt{on})}
    \end{minipage}
    \caption{Multi-view benchmark dataset across kitchen (Sets~1--5, left) and bedroom (Sets~6--10, right) environments rendered in Isaac Sim 5.1.0. In each triplet, from left to right: reference observation ($I_A$), viewpoint shift only ($I_B$), and single intrinsic state change ($I_C$).}
    \label{fig:dataset_all}
\end{figure*}

To rigorously decouple camera viewpoint shifts (extrinsic observation noise) from intrinsic environmental changes, we constructed a 10-set indoor benchmark dataset (five kitchen sets and five bedroom sets) using the Isaac Sim 5.1.0 physics simulator (\texttt{REQ-1.1}). Each set consists of three rendered views ($\text{set}N\_\text{A}$, $\text{set}N\_\text{B}$, and $\text{set}N\_\text{C}$) at a resolution of $1920 \times 1080$ using a wide-angle camera (focal length $14\text{ mm}$, horizontal aperture $20.955$):
\begin{itemize}
    \item \textbf{Reference Observation ($I_A$, $\text{set}N\_\text{A}$):} Captured from Camera~1, depicting the initial scene prior to any environmental state change.
    \item \textbf{Viewpoint-Shifted Static Observation ($I_B$, $\text{set}N\_\text{B}$):} Captured from a nearby viewpoint (Camera~2) within the same room. All object states remain completely identical to $I_A$; only the camera viewpoint has shifted.
    \item \textbf{Viewpoint-Shifted Changed Observation ($I_C$, $\text{set}N\_\text{C}$):} Captured from Camera~2 after applying exactly one intrinsic state change (open/closed or on/off) to a single target object in the scene.
\end{itemize}
Fig.~\ref{fig:dataset_all} shows the rendered multi-view triplets across all ten sets, and Table~\ref{tab:dataset_gt} summarizes the corresponding room environments, ground-truth state changes in $I_C$, and their categories in our change taxonomy (Section~\ref{sec:change_taxonomy}).

\begin{table}[t]
\centering
\caption{Environment Settings and Ground-Truth State Changes in $I_C$ (Sets 1--10)}
\label{tab:dataset_gt}
\resizebox{\columnwidth}{!}{%
\begin{tabular}{clll}
\toprule
\textbf{Set} & \textbf{Room} & \textbf{Intrinsic State Change in $I_C$ (Ground Truth)} & \textbf{Change Category (Section~\ref{sec:change_taxonomy})} \\
\midrule
1 & Kitchen & Refrigerator door (\texttt{closed} $\rightarrow$ \texttt{open}) & State \& Attribute \\
2 & Kitchen & Pot lid (\texttt{lid on} $\rightarrow$ \texttt{lid off}) & Topological \& State \\
3 & Kitchen & Cabinet door (\texttt{closed} $\rightarrow$ \texttt{open}) & State \& Attribute \\
4 & Kitchen & Oven door (\texttt{closed} $\rightarrow$ \texttt{open}) & State \& Attribute \\
5 & Kitchen & Rice cooker lid (\texttt{lid closed} $\rightarrow$ \texttt{lid open}) & State \& Attribute \\
6 & Bedroom & Laptop lid (\texttt{closed} $\rightarrow$ \texttt{open}) & State \& Attribute \\
7 & Bedroom & Cupboard drawer (\texttt{closed} $\rightarrow$ \texttt{open}) & Geometric \& State \\
8 & Bedroom & Cupboard door (\texttt{closed} $\rightarrow$ \texttt{open}) & State \& Attribute \\
9 & Bedroom & Desk drawer (\texttt{closed} $\rightarrow$ \texttt{open}) & Geometric \& State \\
10 & Bedroom & Table lamp (\texttt{off} $\rightarrow$ \texttt{on}) & State \& Attribute \\
\bottomrule
\end{tabular}%
}
\end{table}

\subsubsection{Compared Representations and Controlled Prompt Design}
\label{sec:prompts}
We employed Gemini (model: Hydrogen) as the foundation model. To completely eliminate context leakage across trials, every inference run was executed once in a fresh, independent chat session. We compared two explicit linguistic parameter representations:
\begin{itemize}
    \item \textbf{Baseline Representation (Free-Form Description Only: $o_t^{\text{free}}$, \texttt{REQ-1.4}):} Uses solely the free-form descriptive note $X_1$ generated by applying Prompt~\textbf{P1} (instructing the model to describe the visual scene qualitatively in detail without raw metric coordinates) to each observation $I_X$ ($X \in \{A, B, C\}$).
    \item \textbf{Proposed Representation (Hybrid Linguistic Parameterization: $o_t = o_t^{\text{free}} \oplus o_t^{\text{struct}}$, \texttt{REQ-1.4}--\texttt{1.6}):} Concatenates the free-form note $X_1$ with the structured state list $X_2$ generated by Prompt~\textbf{P2}. Grounded in canonical object-state vocabularies established in external object-state benchmarks such as OSCaR~\cite{nguyen2024oscar} and ChangeIt-Frames~\cite{soucek2022look}, Prompt~\textbf{P2} defines a unified 23-category dictionary covering kitchen and bedroom appliances and fixtures with discrete state slots (\texttt{open}/\texttt{closed}, \texttt{on}/\texttt{off}, \texttt{lid on}/\texttt{lid off}) anchored to static landmarks alongside an explicit \texttt{unknown} tag for occluded or unobservable states, outputting \texttt{(object, state)} tuples.
\end{itemize}
For Scene Change Identification (SCI), raw images were strictly discarded (\texttt{REQ-1.3}) and the text-to-text comparison Prompt~\textbf{P3} (\texttt{REQ-3.1}) was applied across 40 pairs: (1) \textbf{A vs.\ B} ($I_A$ vs.\ $I_B$: viewpoint shift only; ground truth is ``no change'') and (2) \textbf{A vs.\ C} ($I_A$ vs.\ $I_C$: viewpoint shift plus a single intrinsic state change; ground truth is solely the single item in Table~\ref{tab:dataset_gt}).

\subsubsection{Deterministic Evaluation Counting Rules}
\label{sec:counting_rules}
All 40 comparison outputs were audited under identical deterministic counting rules (\texttt{REQ-3.4}) for both representations. Each bullet point reported by the model as a difference or state change was counted as one reported change. Notes explicitly attributed by the model solely to camera viewpoint shifts and lists of unchanged states were excluded, whereas erroneous reports of non-state visual differences (lexical drift) and transitions from a known state to \texttt{unknown} (due to viewpoint shift or occlusion) were counted as false positives.

\subsection{Experimental Results and Quantitative/Qualitative Analysis}
\label{sec:exp_results}
Table~\ref{tab:exp_results} presents the per-set and aggregate quantitative comparison between the baseline and proposed representations across all ten sets.

\begin{table}[t]
\centering
\caption{Quantitative Comparison of Multi-View Scene Change Identification (Sets 1--10)}
\label{tab:exp_results}
\setlength{\tabcolsep}{4pt}
\resizebox{\columnwidth}{!}{%
\begin{tabular}{clcccc}
\toprule
\textbf{Set} & \textbf{Target Object} & \textbf{Baseline A vs.\ B} & \textbf{Baseline A vs.\ C} & \textbf{Proposed A vs.\ B} & \textbf{Proposed A vs.\ C} \\
 & & \textbf{False Pos.\ ($\downarrow$)} & \textbf{Found / False Pos.\ ($\downarrow$)} & \textbf{False Pos.\ ($\downarrow$)} & \textbf{Found / False Pos.\ ($\downarrow$)} \\
\midrule
1 & Refrigerator door & 10 & Yes / 10 & 15 & Yes / 5 \\
2 & Pot lid & 14 & Yes / 14 & \textbf{0} & \textbf{Yes / 0} \\
3 & Cabinet door & 12 & Yes / 10 & \textbf{0} & \textbf{Yes / 0} \\
4 & Oven door & 12 & Yes / 18 & 3 & \textbf{Yes / 0} \\
5 & Rice cooker lid & 14 & Yes / 8 & 10 & Yes / 3 \\
6 & Laptop lid & 15 & Yes / 13 & 1 & Yes / 3 \\
7 & Cupboard drawer & 14 & Yes$^*$ / 15 & 2 & Yes / 3 \\
8 & Cupboard door & 16 & \textbf{No} / 13 & 4 & Yes / 5 \\
9 & Desk drawer & 8 & Yes / 8 & 6 & Yes / 1 \\
10 & Table lamp & 16 & Yes / 15 & 8 & Yes / 2 \\
\midrule
\textbf{Total} & \textbf{All 10 Sets} & \textbf{131} & \textbf{9/10 (90\%) / 124} & \textbf{49 (--62.6\%)} & \textbf{10/10 (100\%) / 22 (--82.3\%)} \\
\bottomrule
\end{tabular}%
}
\vspace{2pt}
\begin{minipage}{\columnwidth}
\footnotesize $^*$In Set~7 Baseline A vs.\ C, the model described the drawer's geometric appearance as a pull-out box without explicitly stating a \texttt{closed} $\rightarrow$ \texttt{open} transition.
\end{minipage}
\end{table}

From the experimental results, three core empirical findings emerge:
\begin{enumerate}
    \item \textbf{100\% Recall of Intrinsic State Changes Under Severe Viewpoint Shifts:} Whereas the free-form baseline failed to detect the true change in Set~8 (Fig.~\ref{fig:dataset_all}h)---misidentifying the two viewpoints as entirely different rooms due to large camera displacement (9/10 sets, 90\% recall)---our proposed hybrid representation preserved spatial identity via static-landmark-anchored canonical descriptions and accurately identified the true state change across all 10 sets (100\% recall) without retaining raw images. Moreover, in Set~7 (Fig.~\ref{fig:dataset_all}g), while the baseline only described the drawer's geometric appearance as a pull-out box without recognizing the semantic state transition, our hybrid schema explicitly captured the \texttt{closed} $\rightarrow$ \texttt{open} transition.
    \item \textbf{72.2\% Overall Reduction in False Positives:} Across all 20 comparisons, total false positives dropped from 255 in the baseline (131 in A vs.\ B and 124 in A vs.\ C) to 71 in our proposed method (49 in A vs.\ B and 22 in A vs.\ C), representing an overall reduction of \textbf{72.2\%} (62.6\% under pure viewpoint shifts in A vs.\ B, and 82.3\% under combined viewpoint and state changes in A vs.\ C). This confirms that the structured state schema $o_t^{\text{struct}}$ (\texttt{REQ-1.5}) effectively suppresses lexical drift in free-form text.
    \item \textbf{Achievement of Zero-False-Positive Inference:} While the baseline failed to achieve zero false positives in any comparison (0/20), our proposed method achieved completely error-free change identification (zero false positives) in two A vs.\ B sets (Sets~2 and~3; Figs.~\ref{fig:dataset_all}b--c) and three A vs.\ C sets (Sets~2, 3, and~4; Figs.~\ref{fig:dataset_all}b--d). Residual false positives in sets with steep camera angle changes (e.g., Set~1 A vs.\ B; Fig.~\ref{fig:dataset_all}a) were primarily triggered by boundary objects leaving the field of view (\texttt{known} $\rightarrow$ \texttt{unknown}), empirically motivating the dedicated uncertainty and occlusion gating module (\texttt{REQ-3.2}) specified for long-horizon deployment.
\end{enumerate}

\subsection{Methodological Justification of the Human-Controlled Protocol as a Proof-of-Concept (v1)}
\label{sec:poc_justification}
In the experiments presented in Sections~\ref{sec:exp_setup} and~\ref{sec:exp_results}, the five core steps marked with \texttt{[!]} in Fig.~\ref{fig:pipeline} and Section~\ref{sec:dialogue_protocol} (\texttt{REQ-1.1}, \texttt{REQ-1.4}, \texttt{REQ-1.5}, \texttt{REQ-1.6}, and \texttt{REQ-3.1}/\texttt{3.4}) were executed under a strictly controlled manual operation protocol by the experimenter. Adopting a human-controlled protocol rather than full programmatic automation at this initial stage establishes rigorous methodological legitimacy as a Proof-of-Concept (PoC) for three fundamental reasons:
\begin{enumerate}
    \item \textbf{Pure Isolation of Representation Bottlenecks (Eliminating Confounding Factors):} In a fully automated end-to-end pipeline, system-level engineering noise---such as API timeouts, asynchronous batch-processing context contamination, or JSON parser syntax errors---can confound experimental outcomes, making it impossible to distinguish intrinsic limitations of the proposed linguistic scene parameterization from pipeline implementation bugs. By manually initiating an independent fresh chat session per trial and strictly controlling identical prompts (\textbf{P1}--\textbf{P3}) and hybrid text concatenation, we purely isolated and measured the intrinsic information preservation and viewpoint invariance of $o_t = o_t^{\text{free}} \oplus o_t^{\text{struct}}$.
    \item \textbf{Full Transparency and Token-Level Auditability:} Because all raw rendered images ($\text{set}N\_\text{A/B/C}$), intermediate generated textual parameters ($\text{A1}$--$\text{C2}$), and comparison reasoning logs ($\text{set}N\_\text{base/prop\_AB/AC}$) are preserved and audited by the experimenter, the exact root causes of failure modes or residual false positives (such as sensitivity to steep camera angle shifts in Set~1 A vs.\ B and \texttt{known} $\rightarrow$ \texttt{unknown} boundary transitions) can be completely traced at the token level without black-box opacity.
    \item \textbf{Establishing an Empirical Gold Standard for the Deterministic Verifier (\texttt{REQ-3.4}):} Auditing all 40 comparison outputs under deterministic counting rules (Section~\ref{sec:counting_rules}) provides the empirical foundation for edge-case decision rules---such as distinguishing visual phrasing variations from genuine state transitions and strictly prohibiting \texttt{COMMIT\_REMOVE} on \texttt{unknown} transitions---that must be encoded into the deterministic rule-based verifier (\texttt{REQ-3.4} and Section~\ref{sec:adaptation}) in subsequent automated phases.
\end{enumerate}

\section{Conclusion}
\label{sec:conclusion}

\subsection{Summary of Contributions}
\label{sec:summary}
In this paper, we formulated long-horizon Textual Scene Memory (TSM) maintenance as \emph{explicit linguistic scene parameterization} and established the formal system requirements specification and pluggable functional architecture for \emph{Textual Scene Mapper (TSM)}, enabling mobile robots in long-term autonomy to maintain a logically consistent environment map under strict raw-image disposal. Our main contributions are summarized as follows:
\begin{enumerate}
    \item \textbf{Spatio-Temporal Dec-POMDP Formulation and Qualitative Spatial Abstraction:} Viewed through the continual learning analogy in SLAM---where irreversible scene-to-text translation updates symbolic scene parameters without experience replay---we formulated long-term map maintenance as a Spatio-Temporal Dec-POMDP. Under the architectural invariant of qualitative spatial abstraction (which eliminates high-precision metric coordinate serialization in favor of intuitive descriptions, e.g., two-significant-digit approximations), we systematized six categories of intrinsic scene changes distinct from extrinsic observation noise alongside a discrete memory action space ($\mathcal{A}_{\text{mem}}$).
    \item \textbf{Complete Modular Requirements Specification (\texttt{REQ-1.1}--\texttt{REQ-3.4}) and Occlusion Decoupling:} We defined strict, implementation-agnostic input--output interface contracts across the entire pipeline, centering on parallel hybrid textual encoding ($o_t = o_t^{\text{free}} \oplus o_t^{\text{struct}}$) anchored to static landmarks with explicit \texttt{unknown} uncertainty tags. By enforcing a structural decoupling invariant between viewpoint-induced occlusion (\texttt{known} $\rightarrow$ \texttt{unknown}) and physical disappearance, our specification prevents catastrophic forgetting of accumulated symbolic parameters without retaining raw images.
    \item \textbf{Foundational PoC Validation of Core Requirements:} Through controlled multi-view experiments in Isaac Sim 5.1.0 (Sets~1--10) that isolated representation invariance from system-level confounding factors while establishing full token-level auditability, we empirically demonstrated that our hybrid linguistic parameterization achieves 100\% recall (10/10 sets) of true state changes while reducing total false positives by 72.2\% (62.6\% under pure viewpoint shifts and 82.3\% under combined viewpoint and state changes) compared to the free-form baseline.
\end{enumerate}

\subsection{Future Work}
\label{sec:future_work}
In future work, we will incrementally instantiate the pluggable extension interfaces specified in Sections~\ref{sec:architecture}--\ref{sec:extension_specs} while preserving the exact input--output contracts validated in v1. First, we will automate the front-end observation trigger (\texttt{REQ-1.2}) via representation-space Bayesian surprise and scale canonical state encoding (\texttt{REQ-1.5}) using external object-state vocabularies (such as OSCaR~\cite{nguyen2024oscar} and ChangeIt-Frames~\cite{soucek2022look}) to evaluate long-horizon patrol trajectories under our automated metrics ($\text{MC-F1}$, $\text{CFR}$, and $\text{IAR}$). Second, we will integrate external scene memory persistence, self-evolving change memory pools (\texttt{REQ-2.1}), spatio-temporal context retrieval (\texttt{REQ-2.2}), and dedicated occlusion gating (\texttt{REQ-3.2}) to further advance continual-learning stability across multi-room, multi-session deployments. Finally, we will execute our two-stage offline policy adaptation contract (\texttt{REQ-2.3})---combining protocol imitation (SFT) with deterministic-verifier-guided constraint optimization (GRPO)---for compact on-device Small Language Models (SLMs), realizing autonomous, real-time, edge-native textual scene memory maintenance without cloud dependency.

\bibliographystyle{unsrt}
\bibliography{ref}

\end{document}